\documentclass{article}
\PassOptionsToPackage{numbers, compress}{natbib}
\usepackage[preprint]{neurips_2026}

\usepackage[utf8]{inputenc}
\usepackage[T1]{fontenc}
\usepackage[normalem]{ulem}
\usepackage[table]{xcolor}
\usepackage{amsfonts}
\usepackage{amsmath}
\usepackage{array}
\usepackage{booktabs}
\usepackage{caption}
\usepackage{fontawesome5}
\usepackage{graphicx}
\usepackage{makecell}
\usepackage{microtype}
\usepackage{nicefrac}
\usepackage{siunitx}
\usepackage{subcaption}
\usepackage{tabularx}
\usepackage{todonotes}
\usepackage{twemojis}
\usepackage{url}
\usepackage{xcolor}
\usepackage{xspace}

\usepackage{etoolbox}
\newrobustcmd\B{\bfseries}
\newrobustcmd\I{\itshape}
\newrobustcmd\G{\color{gray}}
\usepackage{hyperref}
\usepackage{cleveref}

\newcolumntype{G}{>{\columncolor{gray!12}}c}

\newcommand{\NICKNAME}{\textsc{PhysiFormer}}
\newcommand{\nickname}{\NICKNAME\xspace}
\newcommand{\eg}{e.g.\xspace}

\newcommand{\tie}{TIE\xspace}
\newcommand{\oursar}{\ensuremath{\Phi_{\mathrm{AR}}}\xspace}

\makeatletter
\renewcommand{\paragraph}{%
  \@startsection{paragraph}{4}%
  {\z@}{-0.5em}{-0.5em}%
  {\normalfont\normalsize\bfseries}%
}
\makeatother

\title{\nickname: Learning to Simulate Mechanics\\ in World Space}
\author{%
\begin{tabular}{ccc}
Yiming Chen & Yushi Lan & Andrea Vedaldi\\
\end{tabular}\\[0.5em]
Visual Geometry Group, University of Oxford\\[0.5em]
{\small \texttt{\{yiming,yushi,vedaldi\}@robots.ox.ac.uk}}\\
\\
}

\begin{document}

\maketitle
\begin{abstract}
We present \nickname, a diffusion transformer for physically-plausible 3D object motion. Unlike video world models that operate in view-dependent pixel space, \nickname represents objects as 3D meshes expressed in world coordinates. Given the initial vertex positions and velocities, as well as object material type, rigid or elastic, the model samples future vertex trajectories.
While related neural physics approaches build on ad-hoc latent spaces or explicitly enforce rigidity and causality, \nickname shows that excellent results can be obtained without any such inductive biases, by casting vertex trajectory prediction as a single denoising diffusion process directly in world coordinates.
The probabilistic formulation captures uncertainty in the learned dynamics, enabling diverse plausible futures from initial conditions, making this framework potentially useful for applications with unobserved uncertainty. The model features attention factorised over time, space, and objects for efficiency, enabling permutation-invariant multi-object reasoning without needing explicit object encoding. 
Trained on over $100$k simulated trajectories,~\nickname generates rigid and elastic mechanics, and generalises to mixed-material settings, unseen real-world geometries, and larger object counts. It substantially outperforms autoregressive baselines in trajectory accuracy, rigidity preservation, and momentum-based physical consistency.
Our results position coordinate-space diffusion as a promising step toward view-invariant, geometry-aware world modelling for robotics, graphics, and physical design.
Visualisations, code, and models are available at~\href{https://yimingc9.github.io/physiformer}{https://yimingc9.github.io/physiformer}.
\end{abstract}
\section{Introduction}%
\label{sec:intro}
\begin{figure}[t]
\centering
\includegraphics[width=0.95\linewidth]{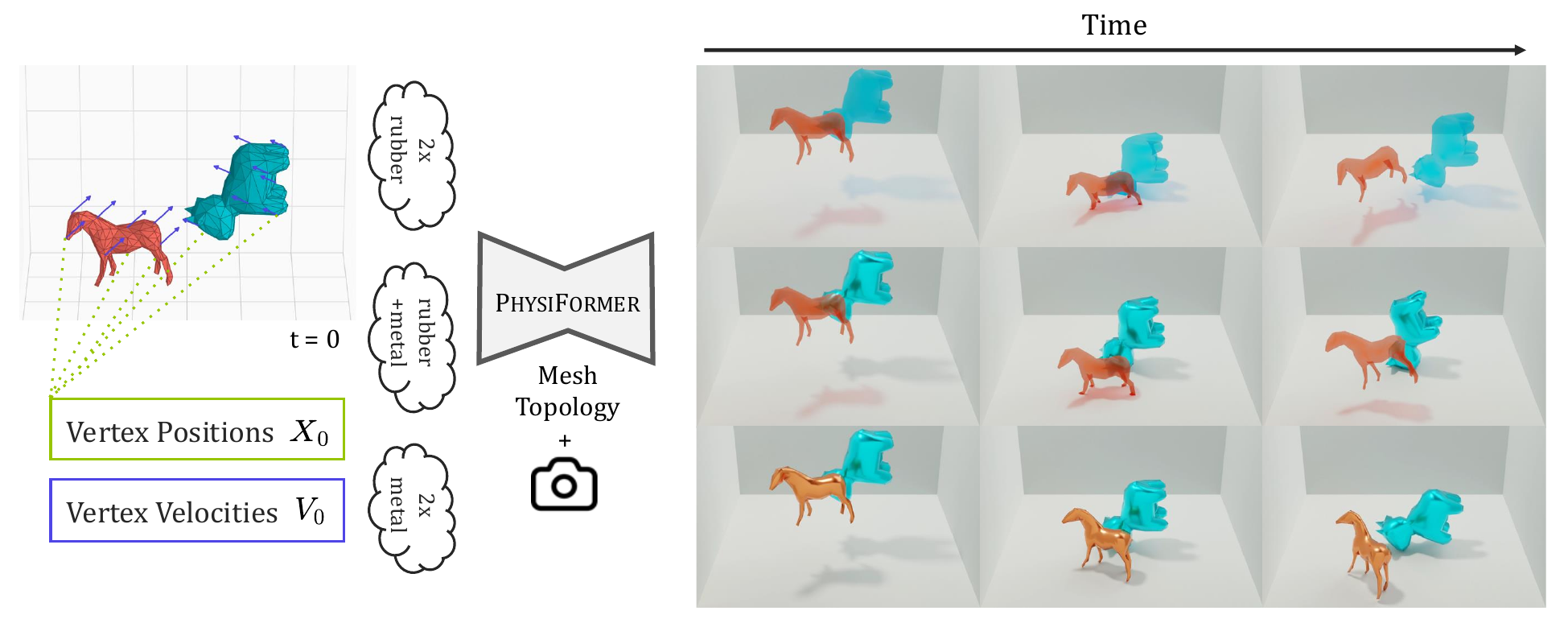}
\caption{\textbf{\nickname{} overview.}
Given initial per-vertex positions $X_0 \in \mathbb{R}^{N \times 3}$ and velocities $V_0 \in \mathbb{R}^{N \times 3}$, and material conditions of (1) rigid, (2) deformable, or (3) mixed, \nickname{} predicts full-sequence future vertex trajectories in a single forward pass, producing physically plausible multi-object dynamics, with mesh topology imposed at inference time.
Output can be rendered as 4D mesh motion under arbitrary conditions.}%
\label{fig:teaser}
\vspace{-6mm}
\end{figure}

Predicting how physical systems evolve over time is central to spatial intelligence and world modelling and has applications to robotics, computer graphics, engineering, and more.
Motivated by large-scale video datasets and the success of diffusion transformer architectures~\cite{peebles23scalable}, several authors~\cite{brooks24video,parker-holder25genie} have sought to reduce the problem of modelling physical systems to video generation.
However, a video is a fairly indirect and ambiguous representation of the state of a physical system, as it intertwines geometry and motion with viewpoint, lighting, and occlusion.

Here we reformulate the problem directly in the space of 3D models, similar to physics simulators like MuJoCo~\cite{todorov12mujoco:} and PyBullet~\cite{coumans16pybullet}.
We are inspired by RenderFormer~\cite{zeng25renderformer:}, which models light transport over triangle meshes, and ask whether a neural network can similarly predict mechanical dynamics using 3D meshes as a representation of the geometry of the system.
We choose 3D meshes because of their ubiquity in applications, making this investigation directly relevant to them.

To answer our question, we introduce \nickname, a diffusion transformer that takes as input 3D mesh vertices representing one or more objects and per-vertex initial velocities.
% with input mesh topology information applied only at inference. 
As shown in \cref{fig:teaser}, \nickname generates the trajectory of mesh vertices over a temporal window, approximating how the system should behave under the laws of mechanics and capturing properties such as inertia, gravity, and collisions.
We also attach different material properties to the objects, allowing them to be either rigid or elastic.

We build our model on top of a general-purpose Diffusion Transformer (DiT)~\cite{peebles23scalable}, conditioned on the initial state of the system to generate the system's future as a sample from an underlying distribution of physically plausible trajectories.
While the architecture is mostly generic, we introduce small but effective modifications to better suit the problem of modelling mechanics.
In particular, we factorise self-attention across time and space to improve the model's efficiency, as well as objects, which implicitly encodes the existence of separate objects in the scene without the need to introduce special tokens or embeddings, and in a permutation-invariant manner.
Unlike recent approaches~\cite{chen26motion}, we perform diffusion in \emph{raw 3D coordinate space}---without the added complexity of learning latent features via autoencoders---following the Just image Transformers (JiT)~\cite{li2025jit} framework.

We train our model on simulated scenes containing either all rigid or all elastic objects undergoing collisions and complex motion.
Once trained, the model generalises to novel object combinations, geometries, dynamic and stationary configurations, mixed-materials, and larger object counts.

A key feature of our model is that it generates the entire trajectory of the system in one go.
This is a departure from the autoregressive (AR) approach often used to model physics~\cite{shao22transformer, sanchez-gonzalez20learning, lin21learning, zhang24dynamic}, where the model is trained to predict the next state of the system given the current state.
AR models are motivated by mechanics, where, under mild assumptions, instantaneous vertex positions and velocities fully characterize the physical state and thereby define a Markovian system.
In fact, authors have applied this AR assumption with some success even when the state is only partially observable in a feature space~\cite{assran25v-jepa,karypidis25dino-foresight:}.

In our experiments, we compare our approach against AR alternatives.
We show that \nickname obtains far better performance.
For one, this is due to a mismatch between training and testing statistics for AR frameworks, which can be mitigated but not removed entirely by Diffusion Forcing~\cite{chen24diffusion} and Self Forcing~\cite{huang25self}.
Moreover, this is due to irreducible error accumulation, which, for instance, causes the shape of rigid objects to gradually deform over time.
In practice, AR modelling works well for particle systems~\cite{shao22transformer, sanchez-gonzalez20learning, lin21learning, zhang24dynamic}, where maintaining the long-term coherence of 3D shape is irrelevant, or in models that hard-code assumptions such as rigidity~\cite{allen22learning, rubanova24learning}, which makes them far less generalisable (for example unable to model elastic objects).

To summarise, our contributions are fourfold:
(1) we introduce a unified neural model that simulates mechanical systems with both rigid and elastic objects directly in 3D trajectory space;
(2) we demonstrate that a general-purpose diffusion transformer with factorised attention over time, space, and objects can model such dynamics effectively;
(3) we show that generative modelling captures uncertainty in properties such as mass and friction which are not provided as conditioning variables, enabling diverse plausible futures impossible for deterministic simulators; and
(4) we find that autoregressive approaches with the same input formulation are limited by error accumulation, or require ad-hoc design choices to compensate for this.
Our findings suggest that spatial intelligence can be supported effectively on top of 3D meshes, which may have direct applications to robotic simulation, gaming, content creation, and engineering.

\section{Related Work}%
\label{sec:related_work}

\paragraph{Traditional Physical Simulation.}

The world is governed by physical laws, and modelling system behavior via physics-based simulation has long been a central goal in engineering.
Substantial progress has been made in motion and material modelling~\cite{muller03particle-based, miguel12data-driven, volino09a-simple}, contact resolution~\cite{bouaziz14projective, li21codimensional}, and time integration~\cite{baraff98large}, spanning discrete and continuum representations for structures such as cloth, thin shells, and fluids.
Despite their strong physical grounding and realism, these approaches remain computationally expensive, algorithmically complex, and difficult to generalize, in part due to domain-specific optimisation constraints.

\paragraph{Per-scene Optimized Physical Dynamics.}

Recent work couples 3D scene representations (\eg, NeRFs~\cite{mildenhall20nerf:} and 3D Gaussian splats~\cite{kerbl233d-gaussian}) with learned dynamics or inverse-physics optimization, typically overfitting per scene to recover deformable object motion from videos~\cite{zhang24dynamic, xie24physgaussian:, zhong24reconstruction, jiang25phystwin:}.
While these methods enable joint reasoning over geometry, appearance, and motion, they often require dense multi-view capture or tracking supervision, and remain limited by strong simulator assumptions (\eg, spring-mass models, simplified contact).

\paragraph{Learning-based Physics Simulation.}

Developments in deep learning have ushered in data-driven approaches to physical simulation.
Convolutional networks have been effective for learning physics on regular grids~\cite{guo16convolutional, bhatnagar19prediction}.
To model more general dynamics, graph neural networks (GNNs)~\cite{scarselli09the-graph} provide a natural framework for particle-based simulation.
Particle states are represented as node features, interactions are encoded by edges, and the system is updated by message-passing conditioned on the previous state~\cite{battaglia16interaction, mrowca18flexible, li19Blearning, sanchez-gonzalez20learning, lin21learning, zhang24dynamic, grigorev23hood:}.
With the advent of Transformers, Transformer with Implicit Edges (\tie{})~\cite{shao22transformer} replace edge-based message passing with attention.
Several works further adapt neural simulation to mesh-based representations:
MeshGraphNets~\cite{pfaff21learning} extend message passing to mesh discretizations, FIGNet~\cite{allen22learning} and HopNet~\cite{wei25integrating} introduce face-based and higher-order interactions for contact-rich dynamics, and HCMT~\cite{yu24learning} combines hierarchical mesh structures with Transformer modules for flexible-body collisions.
This line of work shows the value of mesh-aware inductive biases, but often requires additional geometric machinery such as explicit connectivity, topology preprocessing, hierarchical structures, or learned shape representations as in SDF-Sim~\cite{rubanova24learning}.
In contrast, \nickname models dynamics as diffusion over raw 3D vertex trajectories, supporting multiple objects and materials without specialized contact modeling.

\paragraph{Autoregressive Prediction in Visual Feature Spaces.}

As noted in the previous section, deterministic autoregressive models have become widely adopted for future prediction.
This pipeline can be applied to the Vision Foundation Model (VFM) feature space.
Given four context frames, DINO-Foresight~\cite{karypidis25dino-foresight:} predicts latent futures that can be decoded into interpretable outputs such as depth and segmentation, demonstrating the practical effectiveness of autoregressive prediction.

\paragraph{Diffusion Models for World Simulation.}
 
Video Diffusion Models (VDMs)~\cite{brooks24video, blattmann23stable} are rapidly advancing visual content generation to convincingly simulate the physical world.
However, recent analyses find that VDM outputs frequently violate Newtonian dynamics~\cite{le25what} and do not learn physical laws from videos alone~\cite{kang25how-far-is-video}, limiting their suitability for settings like robotics where physical fidelity is critical.
To address this, several works inject explicit physics signals into VDMs, enabling force-conditioned image-to-video generation with strong generalization~\cite{gillman25force, liu24physgen:, chen2025physgen3d}.
While effective, these approaches require objective-specific training.
Moreover, diffusion in pixel space is inherently view-dependent, which complicates enforcing viewpoint-invariant physical consistency.
Motivated by this limitation, prior work explores diffusion over 3D representations~\cite{Liu2023MeshDiffusion,zhang20233dshape2vecset}, but these methods are largely restricted to generating single, static objects.
More recent efforts add dynamics~\cite{zhang25gaussian, chen26motion}, yet they remain focused on single-object settings and leverage motion in videos for animation.

\section{Method}%
\label{sec:method}

We consider the problem of animating a 3D scene in a physically-plausible manner.
We model the scene as a triangular mesh, with $N$ vertices total, given by positions $X(t) \in \mathbb{R}^{N \times 3}$ indexed by time $t\in\mathbb{R}$, and faces $F \subset \{1,\dots,N\}^3$, defined by triplets of vertex indices.
A scene can contain several objects, modelled as different connected components of the mesh.
We denote the instantaneous velocities of the vertices as $V(t) = dX/dt|_t \in \mathbb{R}^{N\times 3}$.
Given the initial scene configuration, $X(0)$ and $V(0)$, our goal is to draw a sample from a stochastic process $(X(t))_{t > 0}$ that is consistent with the underlying physics of the scene.
We aim to learn a model of this process from data.
%  with the goal of generating physically-plausible animations of the scene from its initial state.

We also assume that, given sufficient initial information, the data can be generated deterministically by a physics simulator (\eg, Genesis~\cite{authors24genesis:}).
This means that there exists a function $\mathcal{S}$ which takes in the state of the system $(X(t), V(t))$ at time $t$ and contextual information $Y$ independent of time to output the future trajectory
$
(X(t'), V(t'))_{t' \geq t} = \mathcal{S}(X(t), V(t)| Y)
$.
Here, $Y$ specifies all properties sufficient to carry out the simulation, such as the mesh topology, gravity, friction and restitution.
% In practice, some parameters (e.g., gravity) are constant, but others differ between scenes.
% In particular,
% In our experiments, we pass to the model some material properties $y \subset Y$ of the scene objects and leave other properties implicit.

Under these assumptions, the dynamics are \emph{Markovian}, which makes autoregressive next-state prediction a natural modelling choice.
However, in practice, it is challenging for a model to learn all relevant state and contextual information, and the Markov assumption is only approximately satisfied.
In our experiments we pass to the model some material properties $y \subset Y$ of the objects, but leave other constant properties implicit, to be determined during learning.

Partially because of this, we find that one-shot trajectory generation is substantially more effective than autoregressive rollout.
Another benefit of such a model is to avoid exposure bias due to the mismatch between teacher-forced training and autoregressive rollout at test time.

\paragraph{Discretizing time.}

To simplify modelling with little loss of generality, we discretize time, and consider trajectories $X, V \in \mathbb{R}^{T\times N \times 3}$ with $T$ steps $t \in \{1,\dots,T\}$.
We use the symbols $X_t = X(t)$ and $V_t = V(t)$ to denote the $t$-th slices of these tensors.
Our goal is to learn the conditional trajectory distribution $p(X \mid X_0, V_0, y)$.
We compare that to autoregressive methods that model the transition
$p(X_{t+1}, V_{t+1} \mid X_t, V_t, y)$ and roll it out iteratively.
Note that autoregressive generation requires predicting both $X$ and $V$ because velocity is part of the Markov state, whereas our one-shot formulation predicts $X$ directly (alternatively, $V_t$ can be approximated as $V_t\propto X_t-X_{t-1}$).

\subsection{\nickname}%
\label{sec:PhysFormer}

\begin{figure}[t]
\centering
\includegraphics[width=\linewidth]{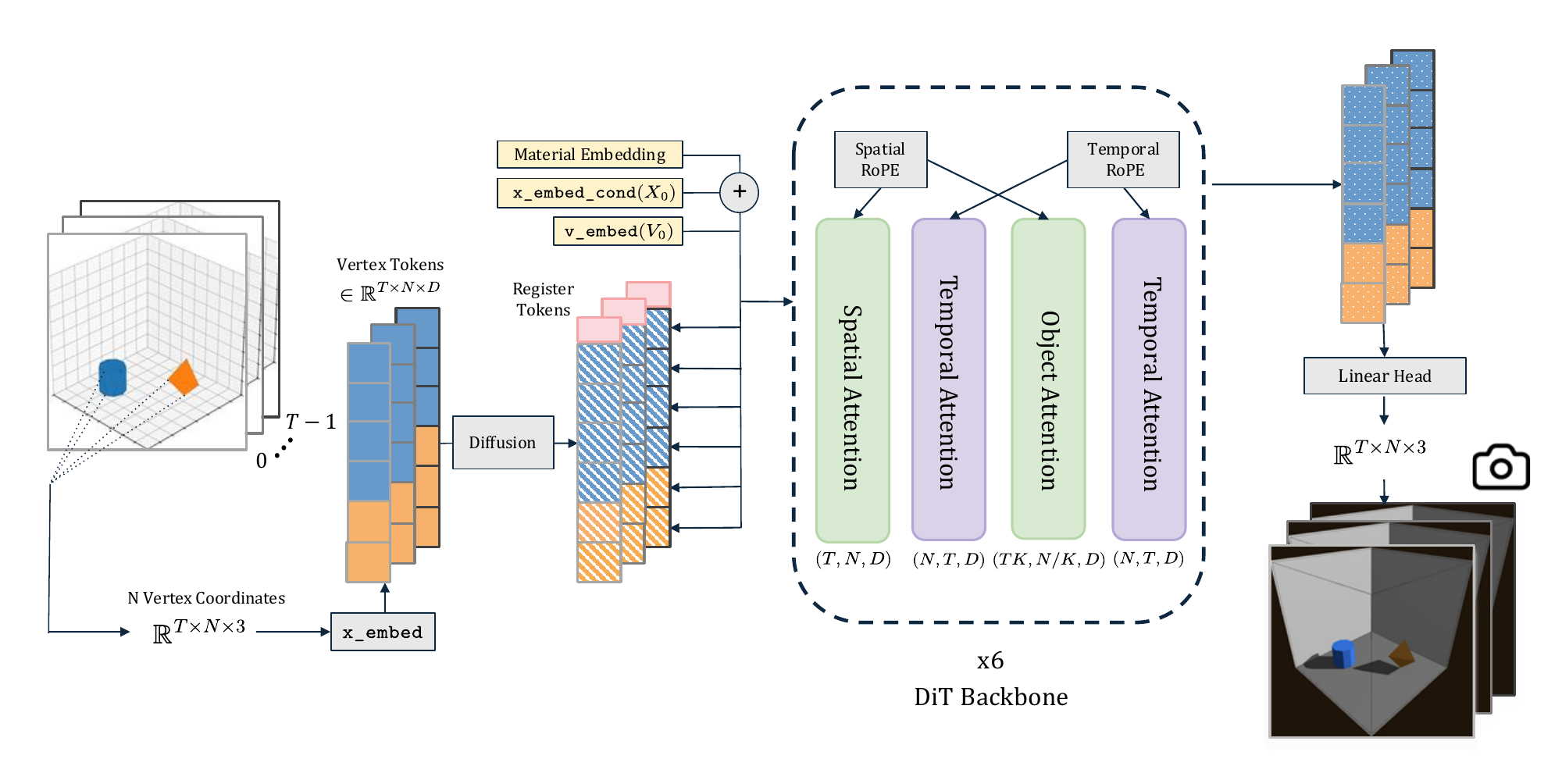}
\caption{
\textbf{\nickname{} Architecture}.
During training, input mesh vertex coordinates in $\mathbb{R}^{T \times N \times 3}$ are projected into hidden dimension $D=1024$ via a linear embedder \texttt{x\_embed}, and diffused with noise according to the flow-matching schedule.
Each noised vertex token is additively conditioned on first-frame position and velocity embeddings (via separate \texttt{x\_embed\_cond} and \texttt{v\_embed}) and a material embedding.
We use 16 prepended global register tokens to aggregate context across the factorized DiT-L backbone with $4\times6$ layers.
The tokens are replicated and consolidated over time, vertices, and objects for spatial, temporal, and object-level attention, respectively, each using its corresponding RoPE\@.
% The resulting token sequence is processed by $12$ pairs of alternating spatial and temporal DiT blocks, each equipped with the respective RoPE.
At inference, iterative denoising produces clean vertex trajectories, which are projected back to coordinate space via a linear head and assembled into triangle meshes using the provided topology for view-invariant, arbitrary-material 4D rendering.}%
\label{fig:physformer}
\end{figure}

In order to model the distribution $p(X|X_0, V_0, y)$, we propose \nickname (\cref{fig:physformer}), a diffusion model that uses a JiT~\cite{li2025jit}-style objective and a Diffusion Transformer (DiT)~\cite{peebles23scalable} backbone.
While we make a point to use a general-purpose architecture, we make minimal modifications to capture the structure of our data.
Instead of introducing object-identity embeddings to distinguish tokens that belong to different objects, we interleave global and per-object spatial attention to make the model object-aware while also being invariant to the order or identity of the objects.
By denoising full trajectories as a single prediction target, \nickname infers positions jointly across time, vertices, and objects, promoting globally consistent generated motion.

\paragraph{Diffusion Model.}%
\label{sec:jit}

We briefly summarise the diffusion framework we use.
Let $x \sim p(x)$ be a random vector from the target data distribution, and let $\epsilon \sim \mathcal{N}(0,I)$ be a normal noise vector of the same dimension.
During training, we construct a noised sample
$
z_\tau = \tau x + (1 - \tau) \epsilon
$
by mixing data and noise according to
$
\tau\in[0,1]
$
{}~\cite{lipman22flow,liu23flow,albergo23building}.
Thus, $z_\tau$ is pure noise for $\tau = 0$ and follows the data distribution for $\tau = 1$.
The flow velocity
$
v(z_\tau,x,\tau)
= {d z_\tau / d\tau}
= (x - z_\tau) / (1 - \tau)
$
is the derivative of $z_\tau$ with respect to $\tau$
(not be confused with the vertex velocity $V$ above).
The model is trained to predict the marginal velocity
$
v(z_\tau,\tau) =
\mathbb{E}_{x,\epsilon} [v(z_\tau,x,\tau) \mid z_\tau].
$
For this, we train a neural network $x_\theta(z_\tau,\tau)$ with parameters $\theta$ expressing the flow velocity as
$
v(z_\tau,\tau) = v(z_\tau,x_\theta(z_\tau,\tau),\tau)
$
and minimising the loss
\begin{equation}
\label{eq:v_loss_jit}
\mathcal{L}(\theta;\tau)
=
\mathbb{E}_{x,\epsilon}
\left \|
v(z_\tau,x_\theta(z_\tau,\tau),\tau) -
v(z_\tau,x,\tau)
\right \|^2
=
\mathbb{E}_{x,\epsilon}
\left \|
\frac
{x_\theta(\tau x + (1 - \tau) \epsilon,\tau) - x}
{1- \tau}
\right \|^2.
\end{equation}
During training, we sample $\tau$ from a logit-normal distribution $\operatorname{logit}(\tau)\sim \mathcal{N}(\mu,\sigma^2)$, where $\mu = -0.8$ and $\sigma = 0.8$.
At inference time, generation proceeds by integrating the corresponding ordinary differential equation (ODE), $d z_\tau/d\tau = v_\theta(z_\tau,\tau)$, from $\tau=0$ to 1 to obtain a sample $z_1 \approx x$.
In practice, we solve the ODE numerically with the Heun integrator with 50 steps.

Because the network outputs the ``clean'' data $x$ while minimising the velocity prediction error, this is called $x$-prediction with $v$-loss.
As suggested by JiT~\cite{li2025jit}, $x$ belongs to a lower-dimensional data manifold than $v$ (see the Manifold Assumption~\cite{chapelle06a-discussion}), which simplifies prediction.

\subsection{Diffusion Architecture}%
\label{sec:diffusion_architecture}

Our goal is to model the distribution $p(X|X_0, V_0, y)$ of vertex trajectories conditioned on initial position and velocity, and contextual information $y$ about the scene.
We thus design a neural network that, given $X_0, V_0, y$ and a noised version $z$ of the data $X$, denoises it to predict the ``clean'' data $X$, i.e.,
$X \approx x_\theta(z_\tau, X_0, V_0, y, \tau)$.
We base our model on general-purpose Diffusion Transformer (DiT)~\cite{peebles23scalable} with modifications that capture the structure of our data, as explained below.

\paragraph{Encoding.}

The transformer requires the data to be converted to a stream of tokens.
To this end, the noisy vertices in $z \in \mathbb{R}^{T\times N \times 3}$ are individually projected to a $D$-dimensional space using a linear layer and the result is flattened to a sequence of $TN$ $D$-dimensional tokens.

The network is conditioned on the initial states $X_0, V_0 \in \mathbb{R}^{N \times 3}$, which are projected to dimension $D$ using separate embedding functions to distinguish initial conditions from model input.
The resulting $N\times D$ tensors are broadcast-summed to the vertex embeddings.

Additionally, we condition the model on two types of object materials by adding material embeddings via a two-layer
$
\mathrm{MLP}_{\mathrm{mat}}
$.
% =
% \mathrm{Linear}_{D \to D}
% \circ
% \mathrm{SiLU}
% \circ
% \mathrm{Linear}_{M \to D}
% $.
Rigid and elastic materials are embedded with inputs 0 and 1 respectively, assigned to objects and broadcast to vertices for efficiency.
The success of this simple conditioning in \nickname suggests its capability to model various physical properties with explicit conditioning in a similar fashion.
The result of the encoder is thus a sequence $z'$ of $TN$ tokens, each of dimension $D$, encoding $z, X_0, V_0$ and the material properties in $y$.

\paragraph{Structured Attention.}

The sequence $z'$ is processed by a stack of DiT blocks.
By themselves, these would operate using self-attention on the entire sequence of length $TN$, with cost $\mathcal{O}(T^2 N^2)$.
We suggest instead structuring attention for efficiency, and also use it to implicitly encode information in the model.
As done in prior works~\cite{blattmann23stable, blattmann23align}, we employ alternating spatio-temporal attention; our novelty is further factorising spatial attention into full spatial and object-level attention.
Concretely, for spatial attention, the $TN$ tokens are reshaped as $(T, N, D)$ treating $T$ as batch dimension, so attention is applied independently within each of the $T$ frames.
For object-level attention, padding the $K$ objects so that they have an equal number of vertices $N/K$, tokens are grouped by object as $(TK, N/K, D)$, 
% where  denotes the maximum per-object vertex count within a batch,
so attention is applied within each object at each time step.
For temporal attention, tokens are reshaped as $(N, T, D)$, so each vertex attends across time.
This factorisation reduces the attention cost to
$
% \mathcal{O}(TN^2 + TN^2/K+ NT^2) = 
\mathcal{O}(TN^2 + NT^2)
$.
% Padding and masking are done appropriately to support batching across variable input lengths.
The interplay between local and global spatial attention makes the model aware of the different objects without using explicit object identifiers and in a manner that is insensitive to the order of the objects in the token sequence.

\paragraph{Spatial and Temporal Rotational Positional Encoding.}

We inject spatio-temporal position information into the transformer blocks with rotary positional encodings (RoPE)~\cite{su24roformer:}, which naturally encode relative information consistent with the global spatio-temporal translation invariance of physical dynamics.
RoPE is applied separately in temporal and spatial attention: temporal attention uses standard 1D RoPE over time indices, while full and object-level spatial attention use coordinate-conditioned RoPE, following RenderFormer~\cite{zeng25renderformer:}, so attention depends on relative 3D vertex offsets.
Specifically, we multiply each of $x,y,z$ by log-spaced base-2 frequencies, concatenate the resulting phases, and convert them to $\sin/\cos$ coefficients for the standard block-wise $2\times2$ RoPE rotations of query and key vectors.
If the number of coordinate-derived phases differs from the per-head rotary dimension, we pad or truncate them; zero-padding gives identity rotations, leaving the remaining channel pairs unrotated.
Sixteen shared register tokens are replicated and consolidated across factorised attention to aggregate global information, with details in \cref{sec:register}.

\vspace{-2mm}
\section{Experiments}%
\label{sec:experiments}

We evaluate the ability of \nickname to simulate the dynamics of rigid and deformable objects.
% In order to do so, we introduce a new synthetic dataset (\cref{sec:dataset}) of such scenes obtained using a physics simulator.
% We then use this dataset to train \nickname, as well as several baselines, and compare their performance across trajectory accuracy and physical-consistency metrics on a held-out test set.
\vspace{-2mm}
\subsection{Dataset}%
\label{sec:dataset}
\begin{table}[b]
\centering
\vspace{-5.5mm}
% \phantomcaption
\caption{Summary of the four datasets used in our experiments.}%
\label{tab:dataset_construction}
\scriptsize
\begin{tabularx}{\linewidth}{lX}
\toprule
$D_1$ &
10k rigid floor-start scenes with 1--5 convex objects from 15 templates ((a) in~\cref{fig:mesh_templates}).
Each object has 4--20 vertices, with at most 88 vertices per scene.
\\
\midrule

$D_2$ &
15k rigid floor-start scenes with 1--5 objects from 25 convex and 10 concave templates ((a) and (b) in~\cref{fig:mesh_templates}). Objects have 4--88 vertices, with at most 356 vertices per scene.
\\
\midrule

$D_3$ &
60k airborne-start rigid scenes: 35k with 1--5 objects and 25k with 6--10
objects. In each object count group, 10k scenes include nonzero initial angular velocity. Objects are selected from the same mesh templates as $D_2$. 
\\
\midrule

$D_4$ &
20k elastic scenes with 1--5 objects. 10k floor-start and 10k airborne-start. Objects are selected from the same mesh templates as $D_2$. 
\\
\bottomrule
\end{tabularx}
\end{table}

We create a synthetic dataset with the Genesis physics simulator~\cite{authors24genesis:}.
Each scene is represented as $(X_0, V_0, X, \mathcal{F}, M)$, where
$X_0,V_0 \in \mathbb{R}^{N\times 3}$ are initial vertex positions and velocities,
$X \in \mathbb{R}^{T\times N\times 3}$ is the ground-truth vertex trajectory,
$\mathcal{F} \in \mathbb{Z}^{N_f\times 3}$ is the triangular face connectivity with $N_f$ faces,
and $M \in \{0,1\}^{N}$ denotes per-vertex material, rigid or elastic.
Objects correspond to connected mesh components with a shared material; the initial velocities are sampled per object and converted to per-vertex velocities; the elastic material is defined to have a fixed Young's modulus. 

We generate 49-frame trajectories in a bounded $[-1,1]^3$ container with timestep $\Delta t = 1/240$s.
Across scenes, we randomize the number of objects, object sizes, shapes, materials, and initial conditions, while keeping density and environmental parameters fixed.
We choose simulator parameters to make collisions as elastic as the simulator allows: friction is minimized, while the simulator’s standard damping is retained for numerical stability. As a result, collisions are near-elastic but not perfectly energy-conserving.

We generate four datasets of increasing complexity, summarized in \cref{tab:dataset_construction}.
Mesh templates used in training are visualised in~\cref{fig:mesh_templates} in the appendix.
All scenes are simulated in the same bounded environment, where objects may collide with each other and the box walls.
In the floor-start rigid settings, objects are placed on the floor without overlap and assigned random initial linear velocities, with zero velocity for a random subset.
A small $2^\circ$ orientation jitter allows initially stationary but unstable objects to fall under gravity.
The airborne-start setting follows the same setup but with objects spawning in the air.
Train/validation/test splits are precomputed using a fixed-seed stratified split to match the data distribution.
All training scenes contain a single material type.

\subsection{\nickname Implementation Details}%
\label{sec:implementation}

\nickname is trained from scratch following JiT's framework in \cref{sec:jit}.
The model predicts vertex positions, which are combined with the original face connectivity for rendering during inference.
We sample noise as $\epsilon\sim\mathcal{N}(0,I)\times$\texttt{noise\_scale}, where \texttt{noise\_scale} = 0.1, and analyze this choice in \cref{sec:ablation}.
Optimization uses AdamW, and we maintain an exponential moving average (EMA) of parameters with decay 0.9999.
Training uses automatic mixed precision (AMP) with bf16.
We use PyTorch Flash SDPA for speed and compatibility with input-dependent masking.
We train on $2$ NVIDIA H100 GPUs with 94GB memory, with an effective batch size of 64 and $lr = 4\mathrm{e}{-5}$.

For comparison with AR baselines, we train a model using the DiT-L backbone on $D_1$ for 70k iterations, denoted as~\nickname-L-10k.
We employ linear learning-rate warm-up over 780 steps, followed by a cosine decay schedule toward $lr = 5\mathrm{e}{-6}$.
We train~\nickname by finetuning~\nickname-L-10k on rigid dynamics in the $D_1 + D_2 + D_3$ rigid dataset for 27k iterations, and then further finetune on elastic object motions in D4 for 12k iterations.
For the latter, we ensure that the model sees a 60/40 ratio of rigid to elastic material scenes.
Inference uses EMA weights, and we clamp first-frame positions to $X_0$ during sampling.
We use $50$ Heun sampling steps for all sampled results in the main paper, though we show that fewer steps yield comparable performance in~\cref{sec:inferencetime}.
%\yc{TODO\@: confirm, trying flexattention to see if faster inference}.

\begin{figure}[t]
\centering
\includegraphics[width=0.9\linewidth]{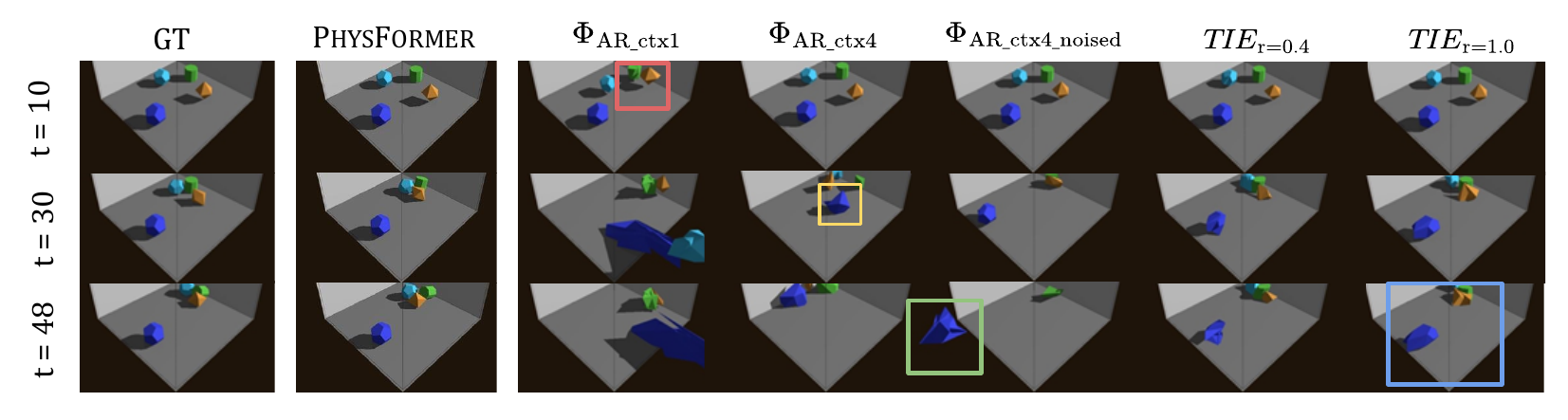}
\caption{
\textbf{Qualitative comparison of \nickname{} against autoregressive baselines on trained 10k rigid object data.}
At $t=10$, \textcolor{red}{\uline{\textcolor{black}{rigidity is not preserved in $\Phi_{\text{AR\_ctx1}}$}}}, but objects remain rigid across all other models.
As $t$ increases, all autoregressive baselines diverge due to error accumulation: \textcolor{yellow}{\uline{\textcolor{black}{stationary objects fail to remain at rest}}}, \textcolor{green}{\uline{\textcolor{black}{objects escape the implicit bounding box}}}, and \textcolor{blue}{\uline{\textcolor{black}{object shapes deform severely, even in the strongest AR baseline \tie$_{r=1.0}$}}}.
\nickname{} consistently maintains object rigidity and produces physically plausible long-horizon motion.}%
\label{fig:baseline}
\vspace{-4mm}
\end{figure}

\subsection{Baselines}

\paragraph{Autoregressive Model \oursar.}

We design and implement a transformer autoregressive framework (\oursar) for next-timestep mesh vertex position prediction.
We optimise performance in two ways inspired by previous works~\cite{chen2024diffusion, karypidis25dino-foresight:}.
First, we explore variable context window lengths.
Longer windows should provide more anchoring information for more stable rollout.
Second, to bridge the train-test domain gap between ground-truth-conditioned training and self-conditioned inference, we investigate noise injection in the context window.
The design choices and implementation details are in the appendix \cref{sec:oursarmethod}.

\paragraph{TIE.}

TIE is an autoregressive \underline{t}ransformer that uses \underline{i}mplicit \underline{e}dges defined by tokens in attention to mimic a graph neural network for next-timestep system-state prediction.
In particle-based dynamics prediction, \tie{} outperforms established GNN baselines~\cite{li19Blearning, sanchez-gonzalez20learning}, continuous convolution methods~\cite{ummenhofer20lagrangian}, and transformer-based models with explicit edge encoding~\cite{dwivedi21a-generalization}.
Our training setup is analogous to a particle-based system, as we define topology only at inference time.
Therefore, we use \tie{} as a strong baseline.
We reproduce \tie{} results using its official training setup, adapted to our mesh trajectory dataset.
Specifically, for fair comparison, we treat each vertex as a separate particle without hard-coding rigidity like in \tie{}'s setup as discussed in \cref{sec:rigidity}.
\tie{} uses radius $r$ to stipulate the maximum distance at which two particles are still connected by an implicit edge captured through attention.
Given the bounding box and range of motion in our dataset, we train the model with $r=0.4$ and $r=1.0$.
We show additional results with larger radii in the appendix \cref{tab:ar_gt_cond}.

\begin{table}[t]
\caption{
\textbf{Comparison of \nickname{}-L-10k against baselines} on 250 test samples, trained on 10k rigid dataset.
MSE measures mean per-vertex position error against GT trajectories;
Rigidity Loss measures implicit rigidity preservation (\Cref{eq:rigidity_error}).
Momentum Drift Ratio measures inference momentum drift from initial system momentum compared against that of GT, where values closer to 1 are better (\Cref{eq:momentum_ratio}).
We show results for 10-frame and 49-frame inference.
Best is \textbf{bolded}; second best is {\itshape italic}.
As \nickname{}-L-10k is a generative model, we show one-shot inference and distributional statistics across 5 generations.
\nickname{}-L-10k is best on average.}%
\label{tab:comparison}
\centering
\small
\setlength{\tabcolsep}{5pt}
\renewcommand{\arraystretch}{1.15}

\begin{tabular*}{\linewidth}{@{\extracolsep{\fill}}lcccccc@{}}
\toprule
& \multicolumn{2}{c}{MSE ($\times10^{-3}$)$\downarrow$}
& \multicolumn{2}{c}{Rigidity Loss ($\times10^{-4}$)$\downarrow$}
& \multicolumn{2}{c}{Momentum Drift Ratio}\\
\cmidrule(lr){2-3} \cmidrule(lr){4-5} \cmidrule(lr){6-7}
Method
& 10-frame
& 49-frame
& 10-frame
& 49-frame
& 10-frame
& 49-frame \\
\midrule
$\nickname{}$ (Ours)
& \textbf{0.0953}
& \textbf{9.55}
& \textbf{0.0411}
& \textbf{0.185}
& 4.45
& \textbf{1.91} \\

\textcolor{gray}{\quad 5 generations $\mu$}
& \textcolor{gray}{0.0883}
& \textcolor{gray}{9.55}
& \textcolor{gray}{0.0407}
& \textcolor{gray}{0.185}
& \textcolor{gray}{4.42}
& \textcolor{gray}{1.94} \\

\textcolor{gray}{\quad 5 generations $\sigma$}
& \textcolor{gray}{0.293}
& \textcolor{gray}{13.5}
& \textcolor{gray}{0.0210}
& \textcolor{gray}{1.22}
& \textcolor{gray}{0.049}
& \textcolor{gray}{0.027} \\

$\Phi_{\text{AR\_ctx1}}$
& 1.78
& 217
& 0.928
& 143
& 21.7
& 11.9 \\

$\Phi_{\text{AR\_ctx4}}$
& 0.896
& 101
& 0.0960
& 27.6
& 3.82
& 8.37 \\

$\Phi_{\text{AR\_ctx4\_noised}}$
& 1.13
& 117
& \textit{0.0846}
& \textit{18.5}
& 3.69
& 7.95 \\

$\tie{}_{r=0.4}$
& \textit{0.157}
& 17.1
& 0.328
& 31.0
& \textit{2.64}
& 2.91 \\

$\tie{}_{r=1.0}$
& 1.73
& \textit{14.8}
& 0.137
& 20.6
& \textbf{2.06}
& \textit{2.29} \\
\bottomrule
\end{tabular*}
\end{table}
\vspace{-2mm}

% % Best-of-5 reports the minimum over five independent samples per test sequence.
% % Dashes indicate deterministic models for which best-of-$k$ sampling is not applicable.
\subsection{Evaluation Metrics}%
\label{sec:evalmetrics}

\paragraph{Trajectory MSE Error.}%
\label{sec:mse}

Following~\cite{shao22transformer, sanchez-gonzalez20learning, ummenhofer20lagrangian, li19Blearning}, given the ground-truth and predicted trajectories $X, \hat{X} \in \mathbb{R}^{T\times N\times 3}$, we define the mean-square error (MSE) as
$
\mathcal{L}_\text{MSE}(\hat{X}|X)
=
\|\hat{X} - X\|_2^2 / (TN).
$

\paragraph{Rigidity Preservation.}%
\label{sec:rigidity}

\newcommand{\rotation}{R}
\newcommand{\translation}{\boldsymbol{b}}

Maintaining object rigidity is a challenge, as seen in severe object deformation across AR baselines in \cref{fig:baseline}.
In \tie{}'s original Boxbath dataset~\cite{li19Blearning}, rigidity is hard-coded via a single predicted rigid transformation for the rigid object (box), which is used to derive motion for all constituent particles.
Thus, we evaluate our model's ability to \textit{implicitly} learn rigidity by averaging the deviation across all objects and frames from a rigid transformation of each object's first-frame position.
We use the Kabsch algorithm~\cite{kabsch76a-solution} to compute a best-fit rigid transform with rotation matrix $\rotation \in SO(3)$ and translation vector $\translation\in\mathbb{R}^3$
% Let $\mathcal{O}_t$ denote the set of objects at time $t$ and let $\mathcal{V}_{t, o}$ denote the set of vertices per object per timestep for $o\in\mathcal{O}_t$
and define the rigidity error as:
\begin{equation}
\label{eq:rigidity_error}
\mathcal{L}_{\text{rigid}}(\hat X | X_0, \mathcal{F})
=
\frac{1}{T}\sum_{t=1}^{T}
\sum_{O \in \operatorname{cc}(\mathcal{F})}
\frac{1}{|O|}
\min_{(\rotation, \translation) \in SE(3)}
\| \hat X_{t,O} - X_{0,O}\rotation - \translation \|_2^\mathcal{F},
\end{equation}
where $\operatorname{cc}(\mathcal{F})$ denotes the connected components of the mesh, $O \subset \{1, \dots, N\}$ is a connected component (object), and $X_{\ast,O}$ denotes the subset of vertices that belong to the component.
% \begin{equation}
% \label{eq:rigidity_error}
% e_{t,o}
% =
% \min_{\rotation
% ,\,\translation
% }
% \frac{1}{3|\mathcal{V}_{t,o}|}
% \sum_{i\in\mathcal{V}_{t,o}}
% \left\lVert
% \rotation\mathbf{v}_{0,i}
% +
% \translation
% -
% \hat{\mathbf{v}}_{t,i}
% \right\rVert_2^2,
% \end{equation}
% where $\mathbf{v}_{0,i}$ is vertex $i$'s position at time 0 and $\hat{\mathbf{v}}_{t,i}$ is the predicted position for vertex $i$ at time $t$.
% The rigidity loss for a single generated sequence is defined as the average alignment residual over all objects and over 48 future timesteps,
% $
% \mathcal{L}_{\text{rigid}}
% =
% \frac{1}{48}
% \sum_{t=1}^{48}
% \frac{1}{|\mathcal{O}_t|}
% \sum_{o\in\mathcal{O}_t}
% e_{t,o}
% $.

\paragraph{Momentum Prediction.}

% The dataset is generated to simulate elastic collisions, but solver limitations fail to conserve momentum exactly.
% Thus, we compare the predicted momentum drift against the ground-truth momentum drift rather than enforcing perfect conservation.

We also test the ability of the model to predict the momentum of the system, which is a fundamental physical property. We compute the momentum of object $O$ at timestep $t$ by approximating the velocity of the center of mass as the average of the vertex velocities computed using finite difference
$
P_{t,O}(X) = m(O) \sum_{i \in O} (X_{t,i} - X_{t-1,i}) / (O\Delta t).
$
We then denote the total momentum at time $t$ as
$
P_t(X) = \sum_{O \in \operatorname{cc}(F)} P_{t,O}(X)
$
and define the momentum drift ratio as:
\begin{equation}
\label{eq:momentum_ratio}
\mathcal{R}_{\text{mom}} (\hat X | X, X_0, F)
=
\left.
\sum_{t=1}^{T}\left \|
P_{t}(\hat X) 
-
P_{0}(\hat X)
\right \|_2^2
~\middle/~
\sum_{t=1}^{T}\left \|
P_{t}(X) 
-
P_{0}(X)
\right \|_2^2
\right.
.
\end{equation}
If the simulated and predicted trajectories are identical, this ratio equals 1.

% We compute object mass $m_o$ using the density specified during data generation and the mesh volume, and define object velocity $\mathbf{u}_{t,o}$ as the finite difference of the center of mass, approximated by the mean of vertex positions, over a timestep $\Delta t$.
% Let $
% \mathbf{p}_{t,o}
% =
% m_o \mathbf{u}_{t,o},
% $ denote the object-level momentum of object $o$ at timestep $t$.
% The total system momentum is defined as the sum of object-level momenta:
% $
% \mathbf{P}_t
% =
% \sum_{o\in\mathcal{O}_t}
% \mathbf{p}_{t,o}
% =
% \sum_{o\in\mathcal{O}_t}
% m_o \mathbf{u}_{t,o}
% $.
% We quantify predicted-versus-ground-truth momentum drift using the ratio $\mathcal{R}_{\mathrm{\mathbf{P}\_drift}}$, for which values closer to 1 are better,
% \begin{equation}
% \mathcal{R}_{\mathrm{\mathbf{P}\_drift}}
% =
% % \frac
% \left.
% {\sum_{t=1}^{48}\left \|
% {\mathbf{P}}^{pred}_t
% -
% \mathbf{P}^{pred}_0
% \right \|_2^2}
% ~\middle/~
% {\sum_{t=1}^{48}\left \|
% {\mathbf{P}}^{GT}_t
% -
% \mathbf{P}^{GT}_0
% \right \|_2^2}
% \right.
% .
% \label{eq:momentum_loss}
% \end{equation}

\subsection{Comparisons Against Baselines}

For fairness, all baselines are trained on the 10k rigid-body dataset $D_1$ using a stratified 9500/250/250 train/validation/test split and are compared with \nickname-L-10k trained on the same data.
Results are shown in \cref{tab:comparison} and \cref{fig:baseline}.
All AR approaches suffer from error accumulation, with mesh deformation and trajectory divergence over time.
Short-horizon autoregressive rollouts over the first 10 steps are meaningfully better.
Across \oursar variants, a longer context window and noise injection help with stability.
For \tie{}, a larger radius models a broader interaction neighborhood, yielding better results.
For the momentum drift ratio, we assume constant mass for computational feasibility.
Because rigidity is largely not preserved in AR baselines, this metric should be interpreted as a rough approximation.
Importantly, we show in the appendix \cref{tab:ar_gt_cond} that all AR models perform well when conditioned on \textit{ground-truth} data at each timestep, highlighting their ability to learn in-distribution dynamics.
Note that \nickname-L-10k displays a large standard deviation for MSE across samples (\cref{tab:comparison}), but MSE alone is not a good indicator of physical plausibility.
Slight contact-angle differences yield different reaction forces in collisions, causing trajectory divergence, while generated rigid motion can differ from ground truth yet remain physically plausible.

\subsection{Generalisation to Unseen Geometries and Object Numbers}

\begin{figure}[t]
\centering
\includegraphics[width=0.95\linewidth]{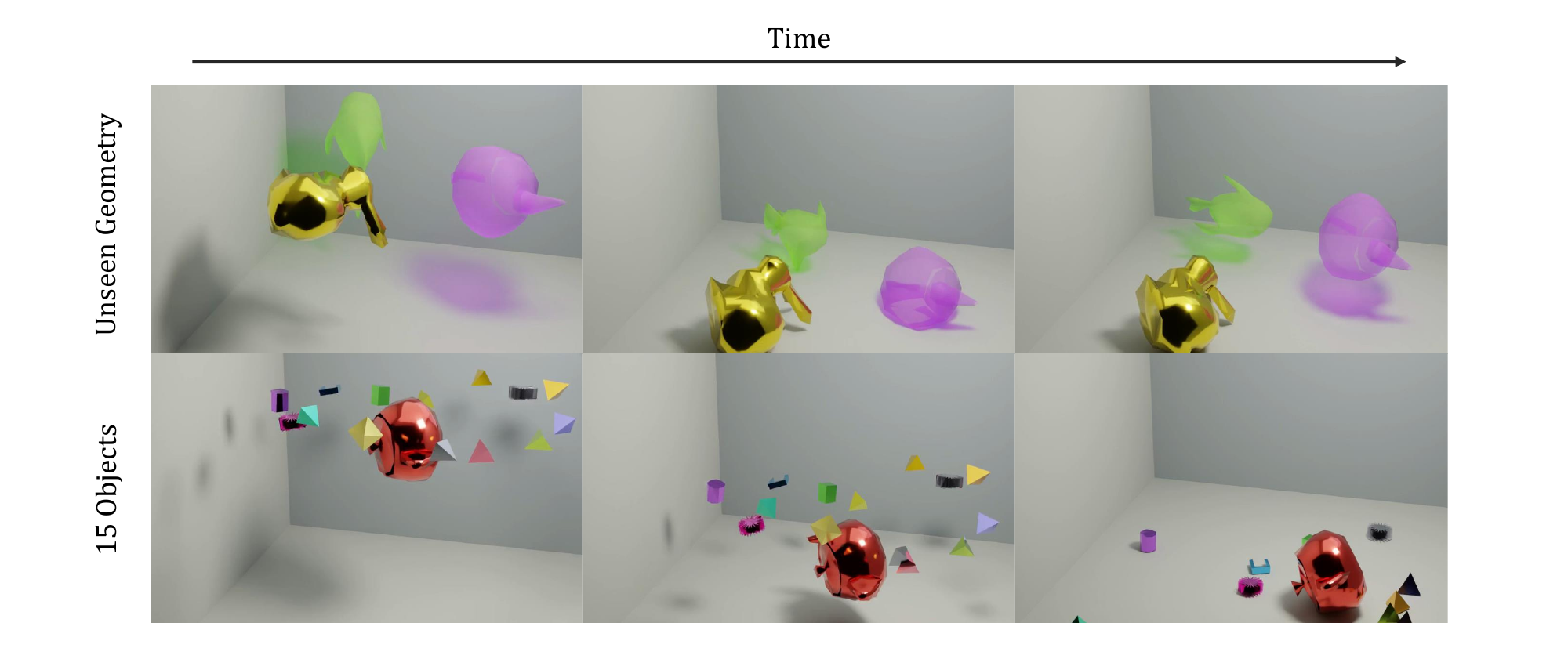}
\vspace{-5pt}
\caption{\textbf{\nickname{} generalizes to complex real-world object geometries and object counts not seen during training}.
Top: Inference on 2 deformable objects (fish and teapot) plus 1 rigid bunny, each with 100 vertices per object.
Deformation is most visible for the middle-frame purple teapot.
\nickname{} allows mixed-material inference although training only saw uniform material across all objects per scene.
Bottom: Inference on 15 rigid objects although training saw at most 10.
Extension to more objects is direct owing to object-level attention, which implicitly encodes object identity in the architecture design.}%
\label{fig:stargen}
\vspace{-4mm}
\end{figure}

\nickname-L-10k generalises well to object geometries and counts not seen in training as shown in the appendix \cref{sec:physformer_gen}.
\nickname is trained on extended data with more complex motion and multiple material properties, and displays impressive generalisation to three settings.
First, it generalises to complex, real-world meshes, with far more vertices per object, although training used only the simpler mesh primitives depicted in (a) and (b) in \cref{fig:mesh_templates}.
Second, it generalises to object counts beyond those seen in training.
Third, it generalises to mixed-material scenes, although training saw only uniform material per scene.
We illustrate these claims in \cref{fig:stargen} and on the project page.

\subsection{\nickname vs. Physics Simulators}

While physics simulators provide high-fidelity trajectories by numerically integrating physical laws, \nickname offers a powerful learned alternative at inference time.
First, it generates physically plausible motion from only initial positions and velocities, without requiring a fully specified physical state such as density, friction, or material parameters as input.
Second, once trained, it has a capped inference cost, enabling efficient rollout even for deformable and contact-rich scenes that are costly to simulate.
On an 80-thread Intel Xeon Gold 6338 CPU node, Genesis rigid-body simulation averaged 1--6.5s per sample for 1--10 objects.
However, elastic-body simulation averaged 20--36s per sample for 1--5 objects excluding rendering, more than 5$\times$ the \nickname inference time on a single H100 GPU for 25 denoising steps, for which we see high-quality \nickname outputs as detailed in~\cref{tab:denoisingsteps}.
Though physics simulators can be sped up, their material-dependent timing gap highlights a key advantage of learned rollout: after training, \nickname{} uses a fixed number of network evaluations rather than costly per-scene simulation.
Third, it generalises to complex, real-world mesh geometries that can be difficult for simulators to handle robustly.
In challenging scenes, where simulators may fail due to contact-resolution artifacts or objects leaving the simulation bounding box as shown in~\cref{fig:leave_box} in the appendix, \nickname can still produce plausible samples.
% \todo{We should provide evidence of these claims}
%rigid data are generated at 90--150 FPS, while elastic data run at only 9--14 FPS\@.

\subsection{Ablations}%
\label{sec:ablation}

\paragraph{Noise Scale in Diffusion.}

Standard diffusion models typically assume normalized inputs, such as VAE latents, and sample noise from a standard Gaussian.
For raw coordinate signals, however, noise scale is important for generation quality~\cite{li2025jit,hoogeboom23simple}.
As shown in \cref{tab:ablations} in the appendix trained on smaller DiT-B based models with 1k data, which we call \nickname-B-1k, we sample $\epsilon \sim \mathcal{N}(0,I) \times \texttt{noise\_scale}$ and find \texttt{noise\_scale} $=0.1$ to work best: smaller values hinder generalisation, while larger values make denoising harder and often introduce jitter.
We hypothesize that this narrower noise distribution is effective because trajectories are strongly conditioned on the first-frame state.
It also stabilizes our coordinate-derived spatial RoPE, whose inputs are noised trajectories during training and early sampling.

\paragraph{Object-Level Attention vs Object-ID Embedding}

We compare \nickname-L-10k's factorised per-object attention with a baseline that uses alternating full spatial and temporal attention plus learnable per-vertex object-ID embeddings in $\mathbb{R}^{(N_{\mathrm{obj}}+1)\times D}$, including a padding ID\@.
On 10k rigid trajectories over 49 frames, the two perform similarly: \nickname has slightly worse MSE (\underline{9.6e-3} vs. 9.1e-3) and momentum drift ratio (\underline{1.70} vs. 1.53), but better rigidity loss (\underline{1.9e-5} vs. 4.8e-5), with comparable qualitative results.
Object-ID embeddings do not extrapolate to more objects, since unseen object indices have no learned embeddings.

\section{Conclusion}%
\label{sec:conclusion}

We presented \nickname, a unified diffusion transformer that learns multi-material, multi-object mechanics as full-trajectory coordinate diffusion in world space.
By generating future mesh motion jointly rather than autoregressively, \nickname avoids error accumulation while preserving object coherence through factorised spatial, temporal, and object-level attention.
It improves trajectory accuracy, rigidity preservation, and momentum-based physical consistency over autoregressive baselines, and generalises to unseen geometries, larger object counts, and mixed-material settings.
Its current limits are fixed trajectory length and mesh resolution, motivating longer-horizon generation, spatial compression, and physics-aware objectives for contact. \nickname uses a general diffusion objective without explicit collision or object-consistency constraints, leading to occasional spurious contacts, interpenetration, and rare orientation discontinuities. These artifacts may be reduced through contact-focused training and tailored physical-consistency losses. Overall, \nickname points toward geometry-level world models for physically plausible 3D dynamics.

\begin{ack}
Yiming Chen is supported by the Rhodes Scholarship.
Yushi Lan and Andrea Vedaldi are  supported by the European Research Council (ERC) grant CoG 101001212-UNION\@.
We thank Isambard-AI and Dawn AIRR supercomputers (project code:  0261--5548--9011--1) for supporting this work\@.%
\end{ack}

\clearpage

\bibliographystyle{plainnat}
% \bibliography{main,vedaldi_general,vedaldi_specific,yslan_general,extra}
% \bibliography{vedaldi_general,vedaldi_specific}
\bibliography{vedaldi_general,vedaldi_specific,yimingc}

\appendix
\clearpage
\begin{center}
{\Large\textbf{\nickname{}: Learning to Simulate Mechanics in World Space}}

\vspace{0.1in}
% {\Large\textbf{\textit{Supplementary Document}}}
{{\textit{Supplementary Material}}}

\vspace{0.1in}

% \vspace{0.1in}
% Anonymous project page: \url{}
\end{center}

% \clearpage
% \section{\nickname{}--DiT backbone details}
% \nickname{} adopts a standard DiT backbone~\cite{peebles23scalable}, and mirrors DiT-L in size, instantiating 24 blocks with 6 sets of 4-block alternation between full spatial, temporal, object-level spatial, and temporal attention blocks. Each block uses 16 attention heads and hidden size $D = 1024$.
% Each block uses RMSNorm~\cite{zhang2019root} and AdaLN~\cite{peebles23scalable} conditioning from the diffusion timestep embedding to produce shift/scale modulation and gated residuals, followed by non-causal multi-head self-attention with QK-normalization~\cite{henry20query-key} and a feed-forward network with SwiGLU activations~\cite{shazeer20glu-variants}.

\section{Method Continued}

\subsection{\nickname{}}%
\label{sec:register}

\paragraph{Register Tokens.}

In input data tokenization, we further prepend $N_{\text{reg}}=16$ shared, learnable register tokens, yielding per-sample token embeddings in $\mathbb{R}^{(N_{\text{reg}}+T\cdot N)\times D}$.
The register tokens capture global context and can potentially remove high-frequency noise in embeddings~\cite{darcet24vision}.
Across the temporal and spatial attention blocks, register tokens are replicated across the corresponding dimensions (time for full spatial blocks, objects and time for object-level spatial blocks, vertices for temporal blocks) before attention.
They are consolidated back to a single set of tokens via averaging after attention, allowing global information to be shared consistently across factorised operations.
Within temporal blocks, temporal RoPE assigns register tokens a fixed time index of 0.
For spatial blocks, each register token uses the mean position of the vertices selected by its context mask for spatial RoPE.

\paragraph{DiT Details.}

Up to the modifications described above, the DiT blocks are standard.
Each block uses RMSNorm~\cite{zhang19root} and AdaLN~\cite{peebles23scalable} conditioning from the diffusion timestep embedding to produce shift/scale modulation and gated residuals, followed by non-causal multi-head self-attention with QK-normalization~\cite{henry20query-key} and a feed-forward network with SwiGLU activations~\cite{shazeer20glu-variants}.
By default, we train \nickname{}, which mirrors DiT-L in size, instantiating 24 blocks in total.
Each block uses 16 attention heads and hidden size $D = 1024$.

\subsection{Baseline \oursar}%
\label{sec:oursarmethod}

\paragraph{Autoregressive Dynamics Modeling.}

The autoregressive predictor \oursar outputs next-timestep velocities for $N$ vertices given context length $L$.
Starting with $L=1$, given inputs $X_t \in \mathbb{R}^{1\times N\times 3}$ and $V_t \in \mathbb{R}^{1\times N\times 3}$, the model predicts next-step velocities $V'_{t+1} \in \mathbb{R}^{1\times N\times 3}$ from the final context-frame velocity token.
As prior work showed modeling relative offsets is more natural for attention than absolute values~\cite{shaw18self-attention}, we predict velocity and recover positions by integration, $X'_{t+1} = X_t + \Delta t\, V'_{t+1}$, where $\Delta t$ is defined during data generation.
During training, we normalize positions and velocities using the dataset-wide global mean and standard deviation computed separately for each quantity.
We supervise on velocity loss $\mathcal{L} = \texttt{SmoothL1}(V'_{t+1}, V_{t+1})$, with $\beta = 1.0$.
At inference, we start from the ground-truth state at $t=0$, predict $V'_{t+1}$, integrate to obtain $P'_{t+1}$, and iteratively roll out by feeding $(P'_{t+1}, V'_{t+1})$ back into \oursar.

% The network \oursar is a transformer-based autoregressive model. During training, we normalize data against dataset global mean and global std, separately for position and velocity. Input sequence length is determined by maximum number of vertices, with masking applied on samples of fewer vertices. We start with context window length $L=1$. Because previous work has illustrated that modeling relative offsets is more natural for attention than absolute values ~\cite{shaw-etal-2018-self}, we predict velocity and integrate over fixed time to obtain absolute position. \oursar takes as input positions $P_t \in \mathbb{R}^{1\times N\times3}$ and velocities $V_t \in \mathbb{R}^{1\times N\times3}$ and outputs next timestep velocities $V'_{t+1} \in \mathbb{R}^{1\times N\times3}$. The next timestep positions are obtained via integration $P'_{t+1} = P_t + \Delta t\times V'_{t+1}$ for constant $\Delta t$ used in data generation. We supervise on velocity loss $\mathcal{L} = \|V'_{t+1} - V_{t+1}\|^2$.

% For autoregressive inference, we start from ground truth $t = 0$ position and velocity, predict $V'_{t+1}$ velocity, integrate to obtain $P'_{t+1}$, and recursively carry out prediction $\Phi_{AR}(V'_{t+1}, P'_{t+1})$.

\paragraph{Context Window.}
Intuitively, a model \oursar that can see more of its past outputs should be able to predict the future with higher consistency.
The context window determines the dependency horizon of the autoregressive framework.
In the spirit of DINO-Foresight~\cite{karypidis25dino-foresight:}, which faithfully predicts next timestep latent DINO features autoregressively with context window 4, we implement $\Phi_{AR\_ctx4}$ with $L=4$ where inputs are $P_{t-4:t} \in \mathbb{R}^{4\times N\times3}$ and $V_{t-4:t} \in \mathbb{R}^{4\times N\times3}$ and output is $V'_{t}$.
At inference, we roll out predictions in a sliding-window fashion, starting with the first 4 ground-truth timesteps.

\paragraph{Train-time Noise Injection.}
For our autoregressive framework on mesh vertex prediction (\oursar), we employ noise injection during training to bridge the train-test domain gap.
In theory, by injecting controlled noise into the context window during training, the model learns to self-correct prediction errors accumulated during autoregressive inference.
We implement $\Phi_{\text{AR\_ctx4\_noised}}$ and carefully tune the noise level to $\epsilon_{pos} \sim \mathcal{N}(0, 0.008^2)$ and $\epsilon_{vel} \sim \mathcal{N}(0, 0.08^2)$ according to the level of error accumulation observed during autoregressive rollout.

\paragraph{\textbf{\oursar} Architecture and Training Details.}
\oursar is a pre-norm Transformer encoder that predicts next-timestep vertex velocities with maximum $N$ vertices, given context length $L$.
We tokenize vertex positions and velocities with shape $L\times N\times 3$, masking samples with fewer vertices.
Each 3D coordinate is encoded with 8 Fourier features and projected to a $C=384$-dimensional token embedding.
Tokens are augmented with a learnable shared base vertex token, temporal embeddings for relative ordering within context window, per-vertex object-ID embeddings, and a type embedding that distinguishes position from velocity tokens.
We prepend $R=4$ learnable register tokens and concatenate all position and velocity tokens.
\oursar has four layers, eight attention heads, a feed-forward dimension of $1024$, and dropout $0.1$.
\oursar{} is trained by enumerating all context window-prediction pairs for all samples (\eg 49--4 = 45 pairs for $L=4$).
We use AdamW optimizer with a constant $lr = 1\mathrm{e}{-3}$ schedule and effective batch size 96.
We train for 300k steps on one NVIDIA H100 GPU with 90GB memory.

\section{Inference Time and Denoising Step Numbers}
% \subsection{Mesh Topologies used in Data Generation}
% For the 10k trajectory dataset, we use the physics simulator Genesis~\cite{authors24genesis:} to generate a diverse, collision-rich dataset.
% We select 1--5 objects from the 15 mesh topologies in \cref{fig:convex_train} with replacement and vary their scales independently.

% We follow a similar protocol when testing out-of-domain generalization with sampling with replacement and independent scaling.
% For unseen convex and concave settings, we select 1--5 objects from the top and bottom sections of \cref{fig:unseen_geom}, respectively; for $>5$ objects we select 6--10 objects from training-set convex objects.

% \paragraph{Per-sample Inference Time.}%
\label{sec:inferencetime}

For reported statistics and visualisations, we apply a 50-step Heun sampler during inference by default. \nickname uses a DiT-L backbone, and we report number of denoising steps against inference time on a single NVIDIA H100 GPU and evaluation metric performance in~\cref{tab:denoisingsteps}. Notably, we observe that fewer denoising steps yields outputs of comparable quality. On 250 test samples from the \nickname-L-10k training dataset, \oursar{} variants follow a ViT-s structure and require 0.2s on a single H100 machine; and \tie{} variants have around 770k parameters and require 0.47s on a single Quadro RTX 6000 machine (chosen for compatibility with legacy Python and CUDA dependencies in the original codebase).
Although \nickname is a larger model, we show that current AR baseline's smaller model can learn in-distribution dynamics~\cref{sec:arcanlearn} with GT-conditioned rollout and attribute failure scenarios in long-horizon autoregressive rollout.
Additionally, we show with \nickname that the larger model can capture significantly more nuanced dynamics on more complex meshes with increased generalisation capabilities.

\begin{table*}[t]
\caption{
\textbf{Physics Simulator Stepping Time vs \nickname{} Inference Time.} 
We analyse the effects of denoising step numbers on evaluation metric performance and inference time on a single NVIDIA H100 GPU. 
Rigid objects (Left): over 350 test samples with 61 vertices on average chosen using randomised stratified selection to reflect data distribution, we observe a linear upward trend for per-sample inference time. While MSE is best at smaller denoising steps, rigidity is best preserved at larger denoising step numbers. Accounting for the extent to which MSE reflects physical plausibility and taken together with qualitative renderings, 25 denoising steps performs well and is comparable to 50 denoising steps. Elastic objects (Right): over 40 test samples with 60 vertices on average chosen using randomised stratified selection to reflect data distribution, per-sample inference time is similar to that of rigid object scenes and exhibits linear increase. MSE and Momentum Drift Ratio are marginally better at 5-10 denoising steps. Qualititative results on the project page show that 5 denoising steps and above appears physically plausible. 
}
\label{tab:denoisingsteps}
\centering
\small
\setlength{\tabcolsep}{3.5pt}
\renewcommand{\arraystretch}{1.15}

\begin{minipage}{0.52\textwidth}
\centering
\begin{tabular}{@{}lGccc@{}}
\toprule
\makecell{Denoising\\Steps}
& \makecell{Inference\\Time}
& MSE
& \makecell{Rigidity\\Loss}
& \makecell{Momentum\\Drift Ratio} \\
\midrule
1  & 0.16s  & \textbf{1.31e-2} & 1.03e-3 & 1.076 \\
5  & 1.2s   & \underline{1.50e-2} & 7.23e-5 & 1.044 \\
10 & 2.5s   & 1.58e-2 & 2.69e-5 & 1.052 \\
25 & 6.4s   & 1.70e-2 & \textbf{1.84e-5} & \textbf{1.021} \\
50 & 12.9s  & 1.79e-2 & \underline{2.02e-5} & \underline{1.034} \\
\bottomrule
\end{tabular}
\end{minipage}
\hfill
\begin{minipage}{0.43\textwidth}
\centering
\begin{tabular}{@{}lGcc@{}}
\toprule
\makecell{Denoising\\Steps}
& \makecell{Inference\\Time}
& MSE
& \makecell{Momentum\\Drift Ratio} \\
\midrule
1  & 0.32s  & \textbf{1.18e-2} & 1.041 \\
5  & 1.36s  & \underline{1.30e-2} & \textbf{1.015} \\
10 & 2.7s   & 1.36e-2 & \underline{1.022} \\
25 & 6.7s   & 1.45e-2 & 1.039 \\
50 & 13.5s  & 1.52e-2 & 1.070 \\
\bottomrule
\end{tabular}
\end{minipage}
\end{table*}

\begin{figure}[t]
\centering
\includegraphics[width=\linewidth]{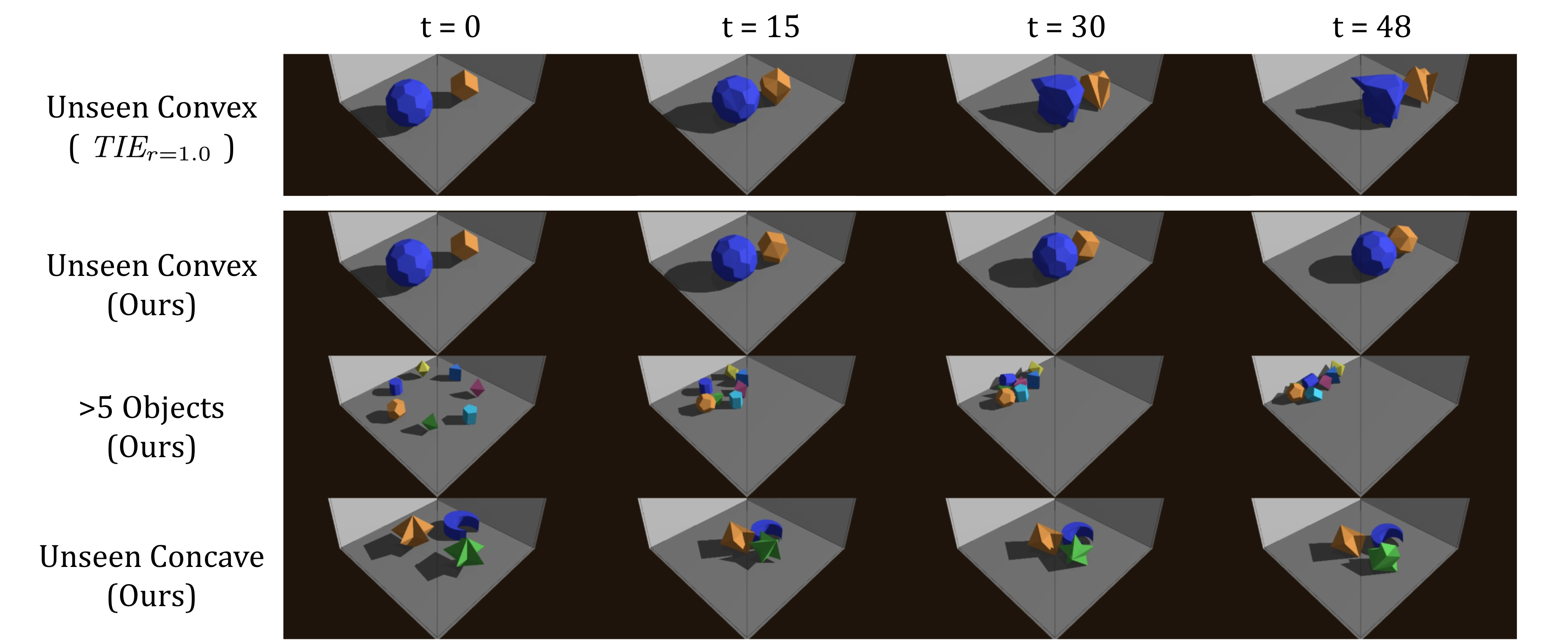}
\caption{\textbf{\nickname{}-L-10k generalizes to object geometries and counts not seen during training}, shown at $t = 0, 15, 30, 48$.
The first row shows the best AR model ($\tie{}_{r=1.0}$) on two unseen convex objects.
For the following rows, we have
\textit{top:} two unseen convex objects,
\textit{middle:} seven objects from seen convex templates, exceeding the training maximum of five,
\textit{bottom:} three objects with unseen concave geometry.
\nickname{}-L-10k produces physically plausible rigid-body dynamics across all settings, demonstrating robustness to topology, vertex count, and scene complexity beyond the training distribution.}%
\label{fig:generalize}
\vspace{-4mm}
\end{figure}

\section{\nickname{}-L-10k Generalization}%
\label{sec:physformer_gen}

\nickname{}-L-10k is trained solely on convex object geometries and 1--5 objects per scene.

As shown in~\cref{fig:generalize} and~\cref{tab:generalization}, \nickname{}-L-10k generalizes well to object geometries and counts not seen in training.
As \nickname{}-L-10k is a generative framework, we additionally include a best-of-5 generation statistic per sample.
For unseen convex shapes, although objects in training have at most 20 vertices, \nickname{}-L-10k is able to generate physically plausible motion for objects with up to 86 vertices, showing our framework's capability to generalize to arbitrary topology.
For object numbers exceeding 5 (maximum seen in training), we see that \nickname{}-L-10k still models interactions between all objects faithfully.

\begin{table}[t]
\caption{
\textbf{\nickname{}-L-10k generalization performance} across in-distribution and out-of-distribution settings on size 10k dataset.
\textit{Training test set}: held-out sequences from the training distribution (1--5 objects, 15 seen convex templates).
\textit{Unseen convex}: 100 trajectories with 1--5 objects from 10 novel convex meshes (6--86 vertices each).
\textit{6--10 obj.\ seen convex}: 100 trajectories with object counts exceeding the training maximum of 5, using seen convex templates.
\textit{Unseen concave}: 100 trajectories with 1--5 objects from 10 novel concave meshes (12--88 vertices each).
Metrics follow~\cref{tab:comparison}.
}%
\label{tab:generalization}
\centering
\small
\setlength{\tabcolsep}{6pt}
\renewcommand{\arraystretch}{1.15}
\begin{tabular}{lcccc}
\toprule
& \multicolumn{2}{c}{MSE$\downarrow$}
& \multicolumn{2}{c}{Rigidity Loss$\downarrow$} \\
\cmidrule(lr){2-3} \cmidrule(lr){4-5}
Setting
& 1 sample
& Best-of-5
& 1 sample
& Best-of-5 \\
\midrule
Training test set       & 9.1e-3          & \textbf{5.5e-3} & \textbf{4.8e-5} & \textbf{5.5e-6} \\
Unseen convex           & 8.6e-3          & 6.3e-3          & 1.3e-4          & 5.7e-5 \\
6--10 obj.\ seen convex & 1.1e-2          & 8.6e-3          & 6.4e-5          & 3.7e-5 \\
Unseen concave          & \textbf{7.3e-3} & 6.1e-3          & 3.3e-4          & 3.1e-4 \\
\bottomrule
\end{tabular}

\end{table}
\section{\nickname{} Training and Inference Meshes}
\begin{figure}[b]
    \centering
    \begin{subfigure}{0.85\linewidth}
        \centering
        \includegraphics[width=\linewidth]{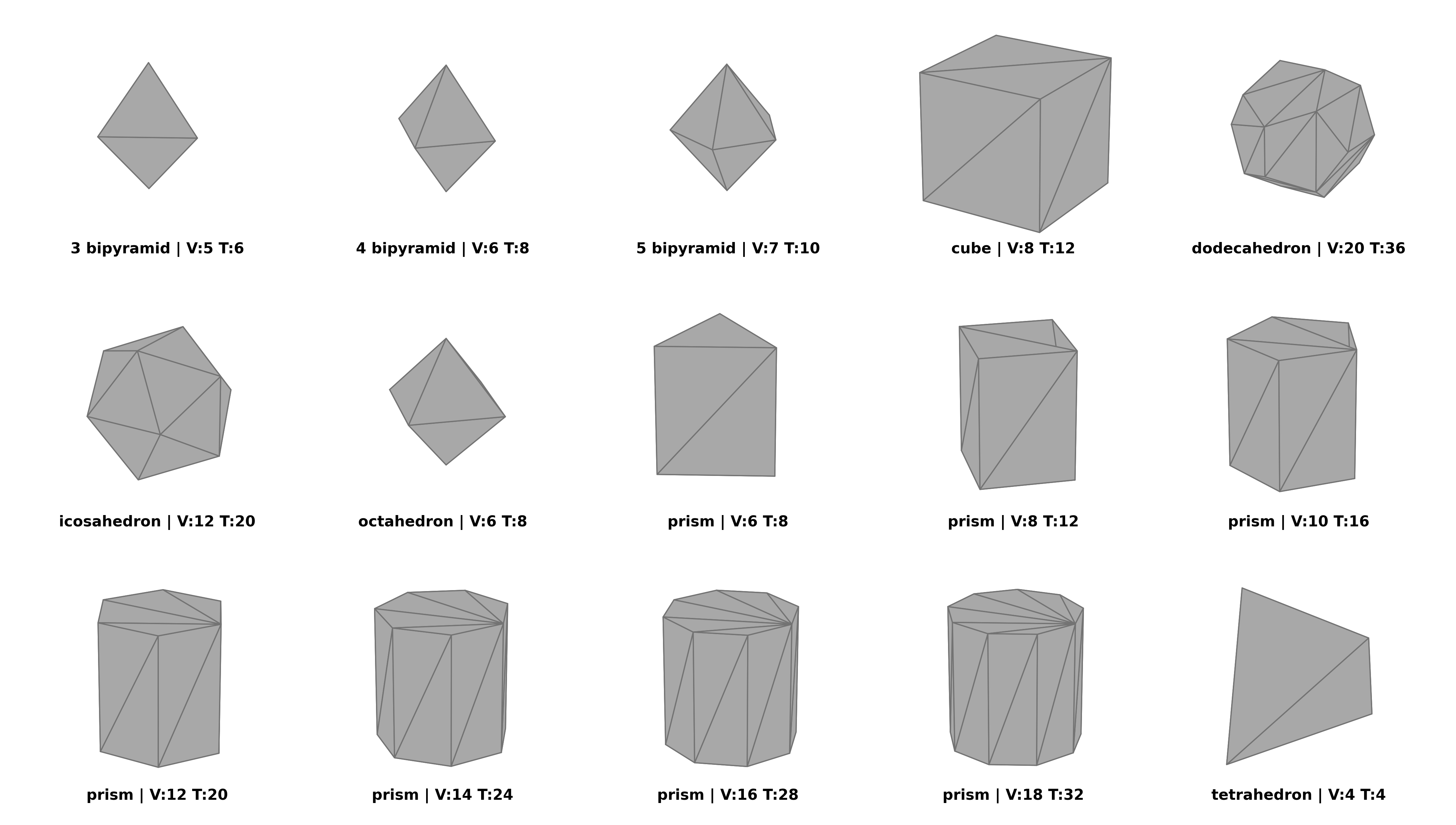}
        \caption{Mesh templates in dataset $D_1$ used to train \nickname{}-L-10k and \nickname{}-B-1k. Each mesh contains 4-20 vertices.}
    \end{subfigure}

    \vspace{0.5em}

    \begin{subfigure}{0.85\linewidth}
        \centering
        \includegraphics[width=\linewidth]{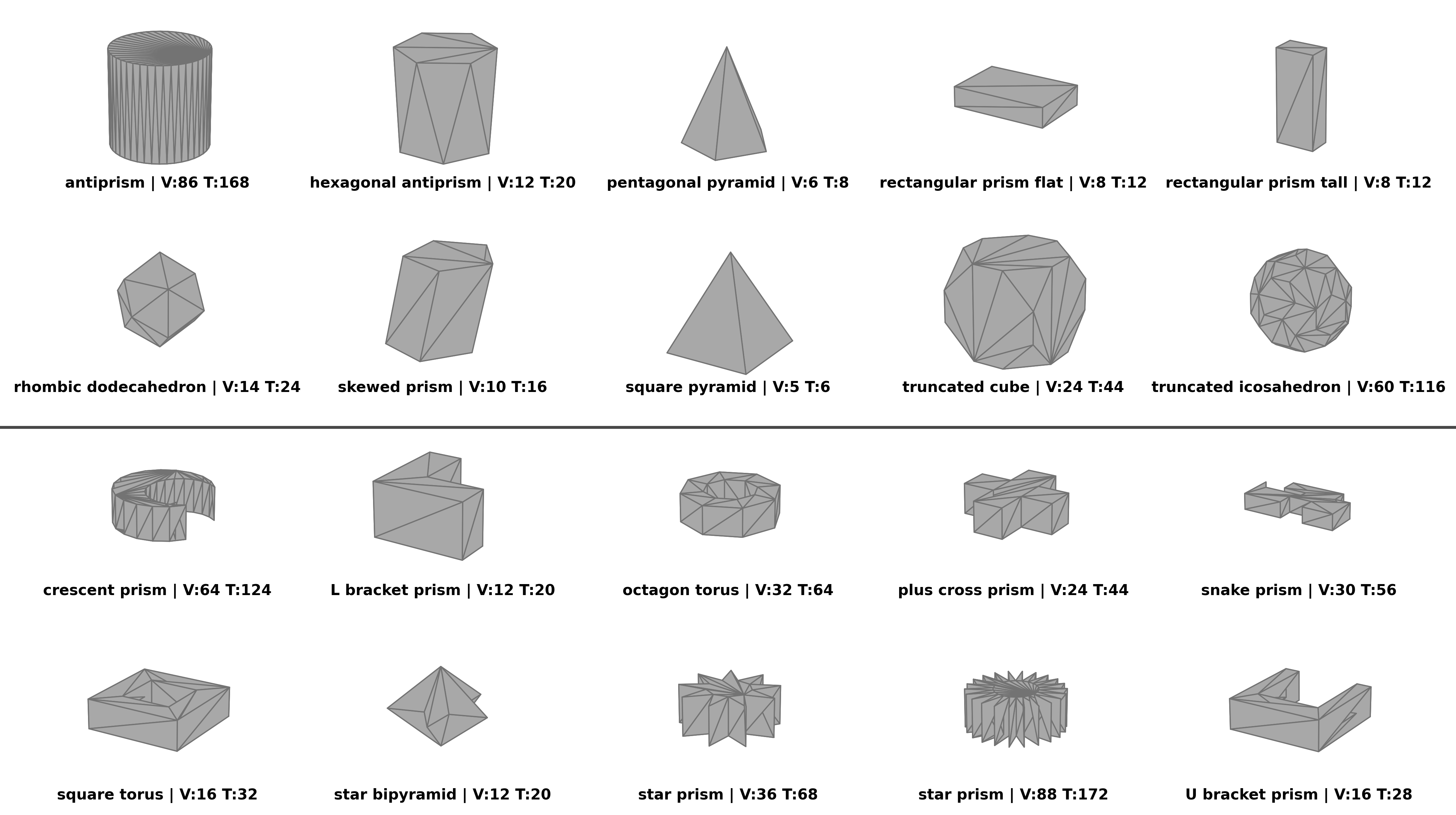}
        \caption{Mesh templates in datasets $D_2$--$D_4$, used together with the templates from $D_1$ to train \nickname. Each mesh contains 4-88 vertices. }
    \end{subfigure}

    \vspace{1.0em}

    \begin{subfigure}{0.8\linewidth}
        \centering
        \includegraphics[width=\linewidth]{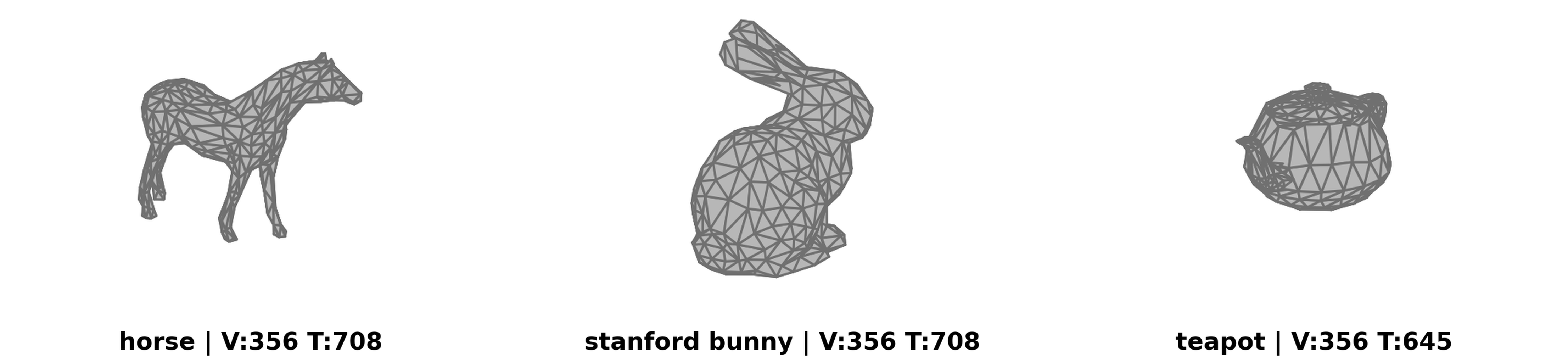}
        \caption{Out-of-distribution complex meshes with far more vertices per object used in \nickname{} inference.}
    \end{subfigure}

    \caption{Mesh templates and real-world geometries used for dataset generation and out-of-distribution inference.}
    \label{fig:mesh_templates}
\end{figure}

(a) in~\cref{fig:mesh_templates} shows mesh templates used to train \nickname{}-L-10k and \nickname{}-B-1k; (a) and (b) shows mesh templates used to train \nickname{}. (c) shows select unseen meshes with significant increase in complexity that still exhibit good results in \nickname{} inference. 

\section{Additional Quantitative Analysis}
\begin{table}[t]
\caption{
\textbf{Comparison between self-conditioned and ground truth-conditioned inference for AR models.}
We average evaluation metrics across 250 test samples from our 10k-trajectory dataset.
For each metric, the first columns show self-conditioned rollout, which follows the true inference-time constraints of autoregressive models, where previous model outputs are used as input for future timestep prediction.
The second columns show ground truth-conditioned rollout, in which the next-timestep prediction uses ground-truth context as input, as during training.
The low error in the ground truth-conditioned setting indicates that the model generalizes well to in-distribution test data and suggests that error accumulation is the primary cause of long-range instability.
The last two rows additionally report results for \tie{} with a larger implicit edge interaction radius, illustrating model performance when accounting for more granular interactions between individual particles.
}%
\label{tab:ar_gt_cond}
\centering
\small
\setlength{\tabcolsep}{6pt}
\renewcommand{\arraystretch}{1.15}
\begin{tabular}{lcccc}
\toprule
& \multicolumn{2}{c}{MSE$\downarrow$}
& \multicolumn{2}{c}{Rigidity Loss$\downarrow$} \\
\cmidrule(lr){2-3} \cmidrule(lr){4-5}
Method
& Self Cond.
& GT Cond.
& Self Cond
& GT Cond. \\
\midrule
 % $\nickname{}$ (Ours) & \textbf{0.0091} & \textbf{0.0055} & \textbf{4.8e-5} & \textbf{5.5e-6} \\
$\Phi_{\text{AR\_ctx1}}$ & 0.22 & 1.9e-5 & 1.4e-2 & 9.8e-7 \\
$\Phi_{\text{AR\_ctx4}}$ & 0.10 & 1.2e-5 & 2.8e-3 & 6.3e-7 \\
$\Phi_{\text{AR\_ctx4\_noised}}$ & 0.12 & 1.2e-5 & 1.9e-3 & 5.8e-7 \\
$\tie{}_{r=0.4}$ & 0.017 & 3.4e-6 & 3.1e-3 & 3.3e-7 \\
$\tie{}_{r=1.0}$ & \textbf{0.015} & 2.9e-6 & 2.8e-3 & 2.5e-7 \\
$\tie{}_{r=2.0}$ & 0.017 & \textbf{2.8e-6} & \textbf{2.0e-3} & \textbf{2.0e-7} \\
$\tie{}_{r=3.5}$ & 0.016 & 2.9e-6 & \textbf{2.0e-3} & 2.1e-7\\
\bottomrule
\end{tabular}
\end{table}

\subsection{AR Models}%
\label{sec:arcanlearn}

Both \oursar{} variants and \tie{} follow an autoregressive inference framework.
In \cref{tab:ar_gt_cond}, we compare self-conditioned autoregressive inference (the true inference setting) with ground truth-conditioned inference (the training setting).
We observe that the latter yields smaller losses and errors, as the model can learn from in-distribution data.
The inference-time failure modes of AR models are therefore attributable to compounding errors during rollout.

Furthermore, for \tie{}, we explored increasing the interaction radius to 2.0 and 3.5.
In a $2\times2\times2$ box, a radius of 2.0 models interactions among most particles, while a radius of 3.5 effectively connects every particle to every other particle via implicit edges.
We omit these radii in the main paper because they offer only a slight advantage on certain metrics, and we aimed to match the radius setting in \tie{}, which connects particles only within a small neighborhood calibrated to the particles' range of motion.
Results are shown in \cref{tab:ar_gt_cond}.

% \subsection{\nickname{} Generation Distribution}
% \label{sec:generationstd}
% As \nickname{} is a generative framework, we show mean and standard deviation across 5 generations  per sample across the test set in~\Cref{tab:comparison} and conclude that it outperforms all baselines on average. 
% Note that for 49-frame generation average, a large standard deviation on the MSE metric sometimes makes \nickname{} worse than TIE baselines, this metric itself is not a good indicator of physical plausibility of output as slight differences in contact collision can lead to dramatically different directions of reaction forces, driving up the error. This number must be viewed in conjunction with other metrics and qualitative output visualizations. 
\begin{figure}[t]
\centering
\includegraphics[width=0.35\linewidth]{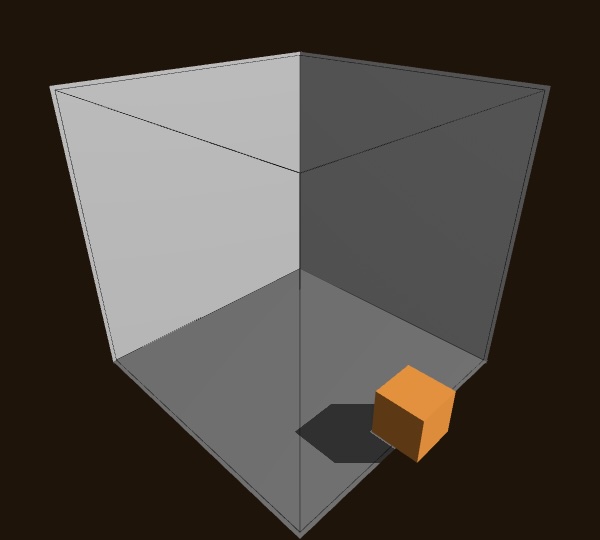}
\vspace{-3pt}
\caption{Physics simulator failure cases occur when boundary contacts are imperfectly resolved, allowing objects to escape the bounding box, especially at high velocities or with fewer simulation substeps.}%
\label{fig:leave_box}
\vspace{-4mm}
\end{figure}
\begin{figure}[t]
\centering
\includegraphics[width=0.9\linewidth]{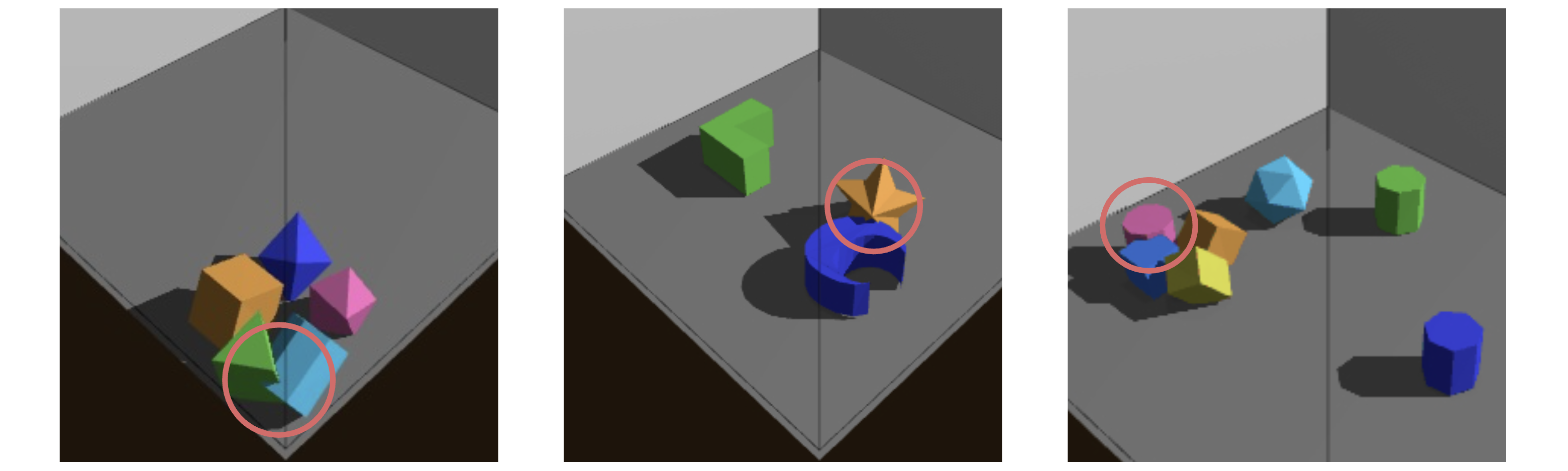}
\caption{Examples of object overlap during inference}%
\label{fig:collision}
\vspace{-4mm}
\end{figure}

\section{Evaluation Metric Discussion}
We follow prior work and use MSE on the ground-truth trajectory for evaluation~\cite{shao22transformer, sanchez-gonzalez20learning, ummenhofer20lagrangian, li19Blearning}.
However, as is apparent from our video visualizations, deviation from the ground truth does not necessarily imply physical implausibility.
In our setting, rigidity loss (when conditioned on rigid materials), momentum drift ratio, and qualitatively visualization assessment paint a more complete picture.
To assess a model's ability to learn physics more fully, improved and more general evaluation metrics are needed in the field.

\subsection{\nickname{} Chunked Long-Horizon Inference}

Current \nickname{} inference output length is determined by the training data length of 49 frames.
Since inference only requires initial position and velocity, we explore \nickname{}'s potential for chunked long-horizon rollout.
For each continuation chunk, the final generated frame of the previous chunk is used as the next initial position.
The next initial velocity is estimated by finite differencing the final two generated frames.
Full-horizon rigidity error increases with rollout length, rising from 6.99e-05 at 1x/49 frames to 6.87e-04 at 4x/193 frames, which yields visible deformations.
This observation motivates further exploration, perhaps via more noise-robust initial-state conditioning in training with noise injection.
Visualizations are available on the project page.

\section{Ablations Continued}%
\label{sec:appen_ablations}

\paragraph{Noise Scale in Diffusion.}

As explained in~\cref{sec:ablation}, the noise scale applied in diffusion training is significant for performance.
Quantitative evaluation results for varying noise scale are shown in~\cref{tab:ablations}.

\paragraph{Mesh Tokenization.}

We also ablate a triangle tokenization analogous to RenderFormer~\cite{zeng25renderformer:}, where each token is the $\mathbb{R}^{3\times3}$ vertex coordinates of a mesh triangle face, flattened to 9D.
In theory, triangle tokens offer a richer local primitive than isolated vertices by implicitly encoding edge geometry, area, and a normal direction, which may simplify learning local surface cues.
However, this induces redundant predictions as each true vertex appears in multiple triangles.
Though we scatter-add and average into one vertex output at inference to ensure connectivity, vertex consistency implicit during training and contributes to the poorer performance in~\cref{tab:ablations}, especially for object rigidity.

\section{Limitations and Future Work}%
\label{sec:limitations}

\nickname{} demonstrates a unified framework for future mechanics prediction across multiple objects and materials.
Currently, due to training dataset specifications, \nickname{} supports generation over 49 frames and performs best with up to 356 vertices.
We aim to incorporate Diffusion Forcing~\cite{chen24diffusion} for autoregressive diffusion to expand the length of generation.
We also want to explore spatial compression with latent encoding~\cite{chen26motion} or learning a latent variational autoencoder like~\cite{zhang233dshape2vecset:} for motion to support even more complex meshes without significantly increasing compute burden.
Finally, \nickname{} is fully data-driven and trained solely with the diffusion loss.
This loss can be augmented with physical inductive biases, such as continuous collision detection (CCD), to mitigate failures like those in \cref{fig:collision}, or with tailored losses to reduce rare object-orientation discontinuities in generated sequences.
\begin{table}[t]
\caption{
\textbf{Ablation study on diffusion noise scale and tokenization strategy.}
All variants use \nickname{}-B-1k (12 total DiT blocks, $D = 768$) trained on a 1k-sample subset, suffixes omitted in table for brevity.
Subscript $ns$ denotes the noise scale used to sample $\epsilon \sim \mathcal{N}(0, I) \times ns$.
\nickname{}$_{\text{tri},ns=0.1}$ replaces per-vertex tokens with per-triangle tokens (flattened 9D coordinates), analogous to RenderFormer~\cite{zeng25renderformer:}.
}%
\label{tab:ablations}
\centering
\small
\setlength{\tabcolsep}{6pt}
\renewcommand{\arraystretch}{1.15}
\begin{tabular}{lcccc}
\toprule
& \multicolumn{2}{c}{MSE $\downarrow$}
& \multicolumn{2}{c}{Rigidity Loss $\downarrow$} \\
\cmidrule(lr){2-3} \cmidrule(lr){4-5}
Method
& 1 sample
& Best-of-5
& 1 sample
& Best-of-5 \\
\midrule
\nickname{}$_{ns=0.05}$         & 0.0073          & 0.0058          & 4.5e-4          & \textbf{1.8e-4} \\
\nickname{}$_{ns=0.1}$          & \textbf{0.0066} & \textbf{0.0056} & \textbf{1.9e-4} & 1.9e-4 \\
\nickname{}$_{ns=0.25}$         & 0.0072          & 0.0069          & 3.2e-4          & 2.7e-4 \\
\nickname{}$_{ns=0.5}$          & 0.0072          & 0.0070          & 2.8e-4          & 2.8e-4 \\
\nickname{}$_{\text{tri},ns=0.1}$ & 0.015         & 0.014           & 1.5e-3          & 1.4e-3 \\
\bottomrule
\end{tabular}
\end{table}

\newpage
\end{document}